\documentclass{article} 
\usepackage{iclr2022_conference,times}

\usepackage{amsmath,amsfonts,bm}

\def\eqref#1{equation~\ref{#1}}

\def\1{\bm{1}}

\DeclareMathAlphabet{\mathsfit}{\encodingdefault}{\sfdefault}{m}{sl}
\SetMathAlphabet{\mathsfit}{bold}{\encodingdefault}{\sfdefault}{bx}{n}

\usepackage{hyperref}
\usepackage{url}
\usepackage{graphicx}
\usepackage{wrapfig}
\usepackage{booktabs}
\usepackage{makecell}
\usepackage{colortbl}
\usepackage{multirow}
\usepackage{fixltx2e}
\usepackage{subcaption}
\usepackage{amsmath}
\usepackage{etoolbox}
\usepackage{multicol}
\usepackage{refcount}
\usepackage{authblk}

\definecolor{LightCyan}{rgb}{0.88,1,1}

\title{UniNet: Unified Architecture Search with Convolution, Transformer, and MLP}

\author{Jihao Liu$^{1,3}$, Hongsheng Li$^{1,2}$, Guanglu Song$^{3}$, Xin Huang$^{3}$, Yu Liu$^{3}$ \\ $^{1}$ CUHK-SenseTime Joint Laboratory \\ $^2$ The Chinese University of Hong Kong \\ $^3$ SenseTime Research}

\iclrfinalcopy 
\begin{document}
\maketitle

\begin{abstract}
Recently, transformer and multi-layer perceptron (MLP) architectures have achieved impressive results on various vision tasks. A few works investigated manually combining those operators to design visual network architectures, and can achieve satisfactory performances to some extent. In this paper, we propose to 
jointly search the optimal combination of convolution, transformer, and MLP for building a series of all-operator network architectures with high performances on visual tasks.
We empirically identify that the widely-used strided convolution or pooling based down-sampling modules become the performance bottlenecks when the operators are combined to form a network. To better tackle the global context captured by the transformer and MLP operators, we propose two novel context-aware down-sampling modules, which can better adapt to the global information encoded by transformer and MLP operators. To this end, we jointly search all operators and down-sampling modules in a unified search space. Notably, Our searched network UniNet (Unified Network) outperforms state-of-the-art pure convolution-based architecture, EfficientNet, and pure transformer-based architecture, Swin-Transformer, on multiple public visual benchmarks, ImageNet classification, COCO object detection, and ADE20K semantic segmentation.  

\end{abstract}

\section{Introduction}
\label{introduction}

Convolutional Neural Networks (CNN) dominate the learning of visual representations and show effectiveness on various downstream tasks, including image classification, object detection, semantic segmentation, etc. 
Recently, convolution-free backbones show impressive performances on image classification \citep{imagenet}. Vision Transformer (ViT) \citep{vit} firstly shows that pure Transformer architecture can attain state-of-the-art performance when trained on large-scale datasets (e.g. ImageNet-21k, JFT-300M). 
Data-efficient image Transformers (DeiT) \citep{deit} is competitive with CNNs when trained with ImageNet only.
MLP-Mixer \citep{mixer} introduced a pure multi-layer perceptron (MLP) architecture that can almost match ViT's performance without using the time-consuming attention mechanism. 
However, the recent transformer or mixer-based architectures are still manually designed with human experience. The different operators (convolution, attention, MLP-mixer, etc.) have different properties, and how to properly assemble them to form networks is still under investigation. The underlying optimal architectures might be quite different for different dataset sizes and computational budgets. On the one hand, convolutions in CNNs are locally connected and their weights are input-independent, which makes it effective at extracting low-level representations and efficient under the low-data regime. On the other hand, the attention operations in the Transformer capture long-range dependency, and the attention weights are dynamically dependent on the input representations. Hence, it requires a significant amount of data and computational resources.
There were recent papers on attempting to manually combine the different types of operators \citep{convit,cvt} to form hybrid convolution-transformer visual networks, which, however, lack proper design principles and did not show promising performance. 
We conducted a pilot study on attempting to directly stack different operators to form networks. 
As shown in Table \ref{pilot_study}, however, the straightforward stacking of different operators achieve even worse result than the original ViT with the only self-attention operator.
In this paper, we search the combination of convolution, transformer, and MLP operators, trying to assemble those operators to create novel and high-performance visual network architectures. 
To the best of our knowledge, we are the first to automatically search the assembly of all types of operators to form novel network architectures. As different operators have distinct characteristics, it is non-trivial to merge them into a super-net that can achieve superior performance. 

\begin{figure}[t]
    \centering
    \begin{subfigure}[b]{0.43\textwidth}
        \includegraphics[width=\textwidth]{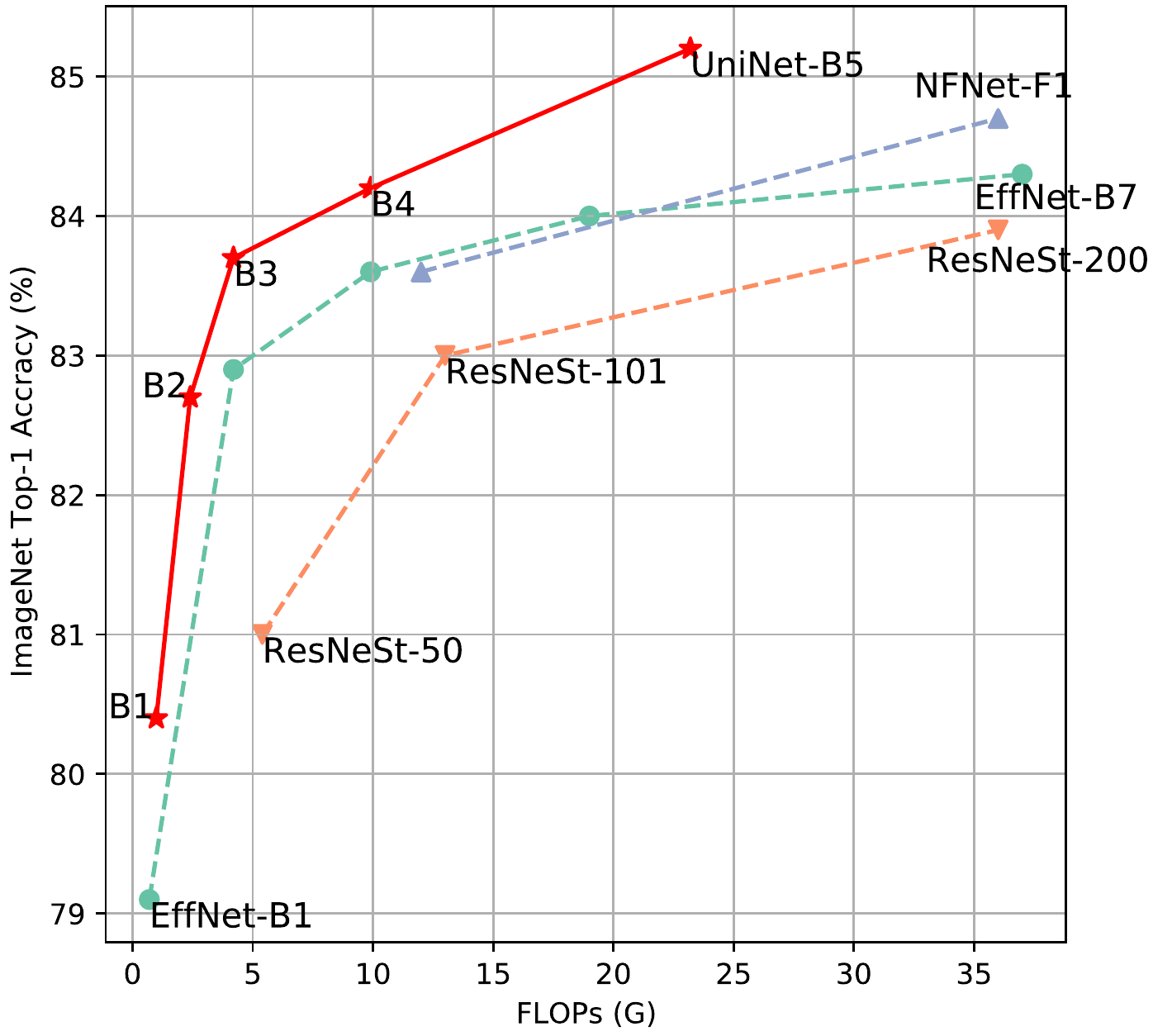}
        \caption{Comparison with ConvNet}
    \end{subfigure}
    \hspace{0.3em}
    \begin{subfigure}[b]{0.43\textwidth}
        \includegraphics[width=\textwidth]{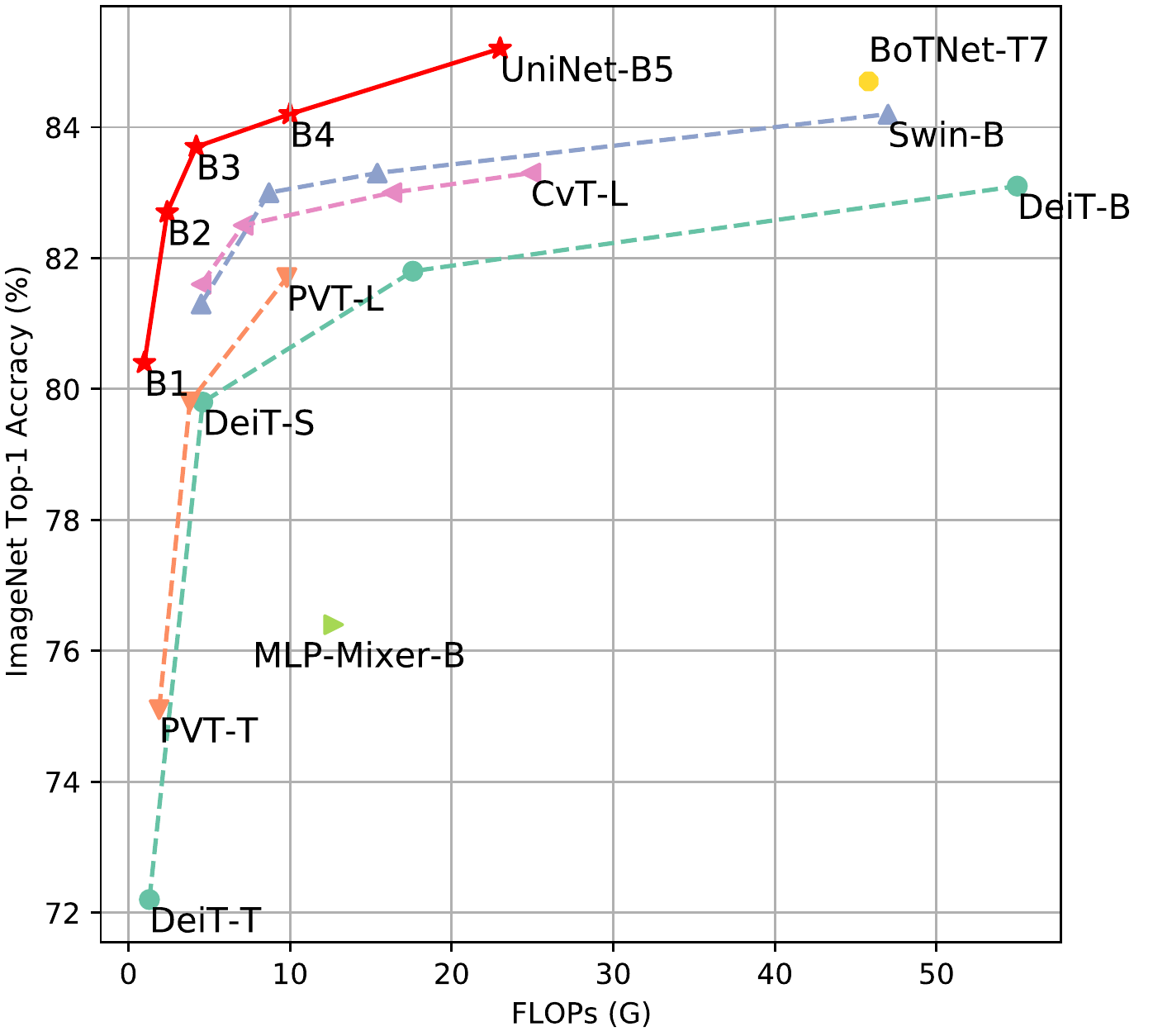}
        \caption{Comparison with transformer or hybrid}
    \end{subfigure}
    \vspace{-1em}
    \caption{\textbf{ImageNet top-1 accuracy vs FLOPs.} All models are trained with ImageNet-1k dataset. Our UniNet-B5 achieve 85.2\% with 23.2G FLOPs, outperforming NFNet-F1 and BoTNet-T7 with 35\% and 49\% fewer FLOPs respectively.}
    \label{fig:sota}
    \vspace{-1em}
\end{figure}

\begin{table}[t]
\centering
\resizebox{0.7\textwidth}{!}{
    \begin{tabular}{c|ccc|c}
    \midrule
    Model & Configuration & \#Params (M) & \#FLOPs (G) & Top-1 Acc. \\ 
    \midrule
    ViT & 12 T & 22 & 4.6 & \textbf{78.0} \\ 
    MLP-Mixer & 18 M & 23 & 4.7 & 76.8 \\ 
    ViT-MLP & 7 T + 7 M & 22 & 4.5  & 76.5 \\
    MLP-ViT   & 7 M + 7 T & 22 & 4.5 & 77.8 \\
    \bottomrule
    \end{tabular}
    }
\vspace{-1em}
\caption{ImageNet top-1 accuracy of different operator combinations. T and M refer to transformer block and mlp-mixer block respectively. Different block numbers are used to keep them comparable.}
\vspace{-1.5em}
\label{pilot_study}
\end{table}

In addition, we find that the widely used down-sampling modules, such as strided convolution or max pooling, actually become the bottlenecks that hinder the searched architectures to achieve optimal performance. This might be derived from the fact that the different operations searched in our architectures, have distinct characteristics. One of the biggest differences is the receptive field. Convolution has local receptive field and is effective at capturing local features, which welly preserve the spatial structure after transformation. However, attention and MLP-mixer have global receptive field, and each pixel in the output feature map is a weighted sum of all spatial locations, which may destroy the spatial structure. Therefore, the locally-correlated down-sampling module is no longer suitable. To address the above issue, we propose three types of down-sampling modules (DSM), which satisfy different operation combinations. Besides the traditional Local-DSM (L-DSM) that are modeled as strided convolution, we propose Local-Global-DSM (LG-DSM) and Global-DSM (G-DSM) to conduct down-sample based on global context. Both LG-DSM and G-DSM use attention mechanisms to gather global features, but the ways to generate query features for calculating the attention weights are different to capture different characteristics of the context. 

We jointly search the operation combination and down-sampling module and network size in a unified search space. We propose a novel scheme to automatically search both the building blocks and the proposed down-sampling modules in a joint manner. 
To find the optimal architecture, we use Reinforcement Learning approach to optimize accuracy and computation cost simultaneously. Our searched architecture, named UniNet (Unified Network), achieves strong results on various vision tasks. 
Our UniNet architectures achieve better performance than state-of-the-art pure convolution architecture, EfficientNet, and pure transformer architecture, Swin Transformer. 
For instance, on ImageNet, our UniNet-B3 is able to achieve 83.7\% and 85.2\% top-1 accuracy with 4.2G and 23.2G FLOPs respectively.
On COCO \citep{coco}, UniNet-B3 achieves 47.9 mAP with 50M parameters, which is also better than the recent local transformer-based architecture Swin-T.

In summary, our contributions are as follows:
\begin{enumerate}
    \item We are the first to jointly search the optimal combination of convolution, transformer, and MLP to identify high-performance visual neural networks.
    \item We reveal the traditional down-sampling module becomes the bottleneck of performance when combining different types of operators. We propose context-aware down-sampling modules, and search them with general processing operations in a joint manner.
    \item Our searched hybrid architecture, UniNet, outperforms previous pure convolution architectures EfficientNet and pure transformer architecture Swin Transformer.
\end{enumerate}

\section{Related Works}
\label{related_works}

\textbf{Convolution, Transformer, and MLP.}
A host of ConvNets have been proposed to push forward the state-of-the-art computer vision approaches such as \citep{resnet,inception,efficientnet}. Despite the numerous CNN models, their basic operations, convolution, are the same. Recently, \citet{vit} proposed a pure transformer based image classification model ViT, which achieves impressive performance on ImageNet benchmark. DeiT \citep{deit} reveals that well-trained ViT can obtain a better performance-speed trade-off than ConvNets. PVT \citep{pvt} proposes a pyramid vision transformer, which can be easily transferred to other downstream tasks. On the other hand, recent papers are attempting to use only MLP as the building block. MLP-Mixer \citep{mixer} and ResMLP \citep{resmlp} show that a pure MLP architecture can also achieve near state-of-the-art performance. 

\textbf{Combination of different operators.}
Besides, another line of works tries to combine different operators to form a new network. CvT \citep{cvt} propose to incorporate self-attention and convolution by generating $\mathtt{Q}$, $\mathtt{K}$, and $\mathtt{V}$ in self-attention with convolution. CeiT \citep{yuan2021incorporating} replace the original patchy stem with convolutional stem and add depthwise convolution to FFN layer, which obtains fast convergence and better performance. ConViT \citep{convit} tries to unify convolution and self-attention with gated positional self-attention and is more sample-efficient than self-attention. Previous works use a different form to combine those operators and get promising results, but requires tedious manual design, lacking effective guidelines. In this work, we propose a unified search space, in which we can search the combination of different operations automatically.

\textbf{Down-sampling module.}
In ConvNets, the down-sampling module (DSM) is implemented with strided-Conv or pooling.  As DSM breaks the shift invariant of convolution, \citet{zhang2019shiftinvar} propose anti-aliased DSM to keep it. Besides, a line of works tries to preserve more information when down-sampling with a learnable or dynamic kernel \citep{lip,dpp,carafe++}. Most of their approaches are down-sampling based on local context, which we show is not suitable for our unified network. In our work, we propose context-aware DSM and jointly search with operation combinations, which guarantees better performance.

\section{Method}
\label{method}

\subsection{Unified Architecture Search}
\label{sec:uas}

As discussed in previous works \citep{convit}, an appropriate combination of convolution and transformer operators can lead to performance improvements. 
However, the previous approaches \citep{cvt, yuan2021incorporating} only adopt convolution in self-attention or feed-forward network (FFN) sub-layers and stack them repeatedly. Their approaches did not fully explore their combinations to take advantage of their different characteristics.
Besides, effective guidelines of properly assembling those operators are missing in prior works, which is important for designing new architectures.

\begin{figure}[t]
    \centering
    \includegraphics[width=0.8\textwidth]{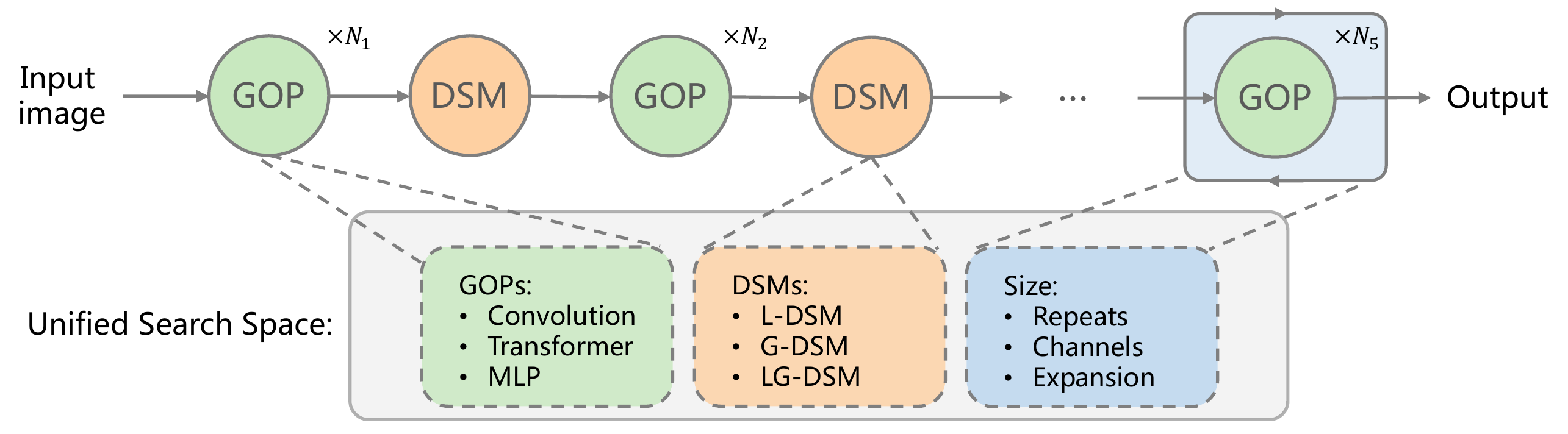}
    \vspace{-0.5em}
    \caption{\textbf{Unified Architecture Search.} We jointly search the all operators and down-sampling modules and network size in a unified search space.}
    \label{fig:backbone}
    \vspace{-1.5em}
\end{figure}

Prior art \citep{carafe++, zhang2019shiftinvar} show that the down-sampling module plays an important role in visual tasks. Most previous approaches adopt hand-crafted downsampling operations, i.e. strided convolution, max-pooling, or avg-pooling, to down-sample the feature map based on only the local context. However, these operations are specifically designed for ConvNets, and might not be suitable to the transformer or MLP based architectures, which capture representation globally. 

In this paper, we jointly search the combination of convolution, transformer, and MLP operators, trying to assemble the operators to create novel and high-performance visual network architectures. For better transform features across different operator blocks, we proposed context-aware down-sampling modules. We jointly search the operators, down-sampling modules, and network size in a unified search space. In contrast, previous Neural Architecture Search (NAS) works achieved state-of-the-art performances mainly via searching the network sizes. We show that our proposed unified architecture search can achieve very promising performance with optimal all-operator network architectures.

In the remaining parts of the section, we firstly present how to properly combine different operators into a unified search space and search them jointly. We then present the challenge of incorporating down-sampling modules with different operators, and present our proposed context-aware down-sampling module. Finally, we will introduce our UniNet architecture and NAS pipeline.

\subsection{Modeling Convolution, Attention, MLP with a Unified Searchable Form}
\label{sec3.2}

Recently, transformer and MLP based architectures are able to achieve comparable performance with convolution-based ones on general visual tasks. To achieve better performance, it is intuitive to assemble all the types of operators to build high-performance all-operator networks.
Actually, a few works \citep{cvt, yuan2021incorporating, convit} have been studied to empirically combine convolution and self-attention. However, manually search of network architectures is quite time-consuming and cannot ensure optimal performances with different computational budgets.

We introduce a unified search space that contains General Operators (GOPs, including convolution, transformer, and MLP), and then search for the optimal combination of those operators jointly. Compared with the prior art, we use a unified form to characterize each operator. Specifically, we use inverted residual \citep{mbv2} to model a general operator block, which first expands the input channel $\mathtt{c}$ to a larger size $\mathtt{ec}$, and later projects the $\mathtt{ec}$ channels back to $\mathtt{c}$ for residual connection. The $\mathtt{e}$ is denoted as the expansion ratio, which is usually a small integer number, e.g., 4. The general operator block is modeled as
\begin{equation}
    \mathtt{y} = \mathtt{x + Operation(x)},
\end{equation}
where $\mathtt{Operation}$ can be convolution, MLP, or self-attention operators, and $\mathtt{x, y}$ represent input and output features, respectively. For convolution, we place the convolution operation inside the bottleneck \citep{mbv2}, which can be expressed as
\begin{equation}
    \mathtt{Operation(x) = Proj_{ec\rightarrow c}(Conv(Proj_{c\rightarrow ec}(x))).}
\end{equation}
The $\mathtt{Conv}$ operation can be either regular convolution or depth-wise convolution ($\mathtt{DWConv}$) \citep{xception}, and the $\mathtt{Proj}$ represents a linear projection.
For self-attention and MLP, operating on the large bottleneck feature map can be quilt slow. Following previous works \citep{vit, mixer}, we separate them from the bottleneck for computation efficiency, and the $\mathtt{Proj}$ is implemented inside the FFN \citep{attention} sub-layer. Each transformer block has a query-key-value attention sub-layer and an FFN sub-layer. For MLP block, we implement with transpose-FFN-transpose like \citet{mixer}.
\begin{gather}
    \mathtt{y = y' + FFN(y')}, \\
    \mathtt{\textrm{where} \quad y' = x + SA(x) \; \textrm{or} \; x + MLP(x), \;  FFN(y') = Proj_{ec \rightarrow c}(Proj_{c\rightarrow ec}(y')),}
\end{gather}
where $\mathtt{SA}$ can be either vanilla self-attention or local self-attention $\mathtt{LSA}$, and $\mathtt{MLP}$ refers to token-mixing operation.

There are two main advantages of representing the different operators in a unified format and search space: (1) We can characterize each operator with the same set of configuration hyper-parameters except for the operation type, e.g. expansion rate and channel size. As a result, the search space for the operator is much simplified, which can speed up the search process. (2) Under the same network size configuration, each operator block has a similar computation cost. The comparison between different operator combinations is fairer, which is extremely important for NAS \citep{mnas}.

\subsection{Context-Aware Down-Sampling Modules}
\label{sec:dsm}

As discussed in Section \ref{sec:uas}, the down-sampling module (DSM) plays an important role in visual tasks. In addition to hand-crafted DSM (i.e. max-pooling or avg-pooling), a few works \citep{dpp,lip,carafe++} tried to preserve more information when down-sampling with learnable or dynamic kernel. Most of their approaches are down-sampling based on local context, which suits for ConvNets well. However, in our unified search space, operators with different receptive filed can be assembled unrestrictedly to form a novel architecture, where the local context may be destroyed and therefore the previous approaches will fail. 

\vspace{-1em}
\begin{figure}[h]
    \centering
     \begin{subfigure}[b]{0.15\textwidth}
         \centering
         \includegraphics[width=\textwidth]{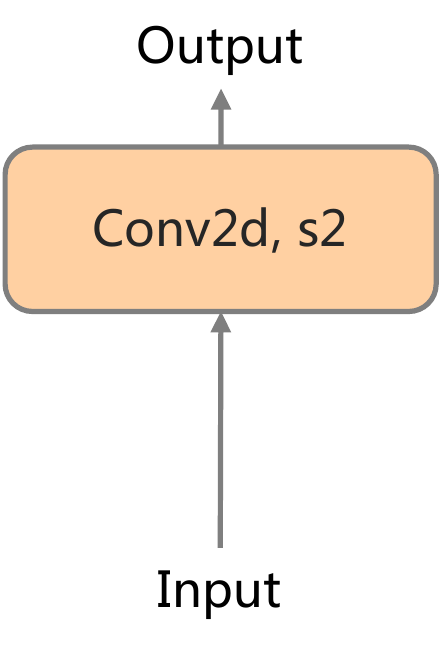}
         \caption{L-DSM}
         \label{fig:ldsm}
     \end{subfigure}
     \hspace{3em} 
     \begin{subfigure}[b]{0.15\textwidth}
         \centering
         \includegraphics[width=\textwidth]{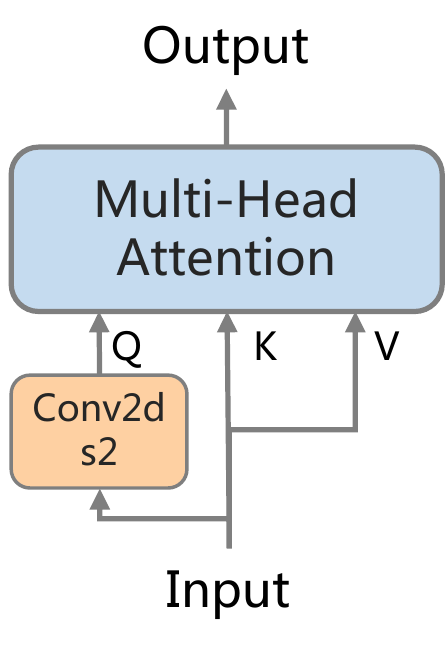}
         \caption{LG-DSM}
         \label{fig:lgdsm}
     \end{subfigure}
     \hspace{3em} 
     \begin{subfigure}[b]{0.15\textwidth}
         \centering
         \includegraphics[width=\textwidth]{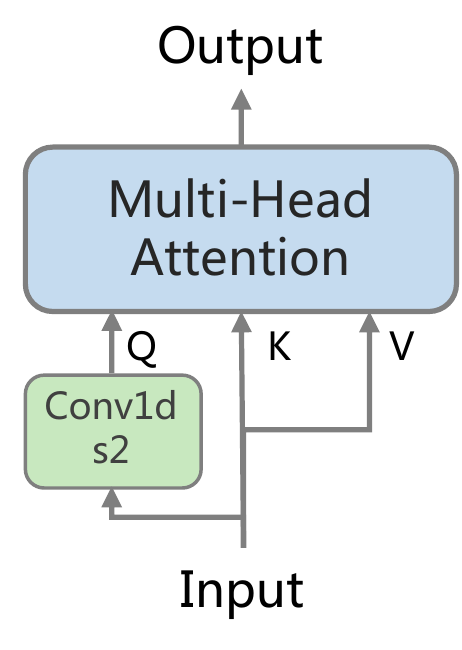}
         \caption{G-DSM}
         \label{fig:gdsm}
     \end{subfigure}
    \vspace{-0.5em}
    \caption{Structures of the context-aware down-sampling modules.}
    \label{fig:dsms}
   \vspace{-1.5em}
\end{figure}

In this paper, we propose context-aware DSM, which is instanced with L-DSM, LG-DSM, and G-DSM. The main difference between those DSMs is the context used when down-sampling. For L-DSM, only local context is involved, which fits ConvNets well as shown in previous works. For G-DSM, only global context is used to down-sample, which may fit other operations, e.g., transformer. The LG-DSM combines the characteristics of L-DSM and G-DSM. It uses both local and global context to downsample. Our intuition is that one of the largest dissimilarities of different operators is the receptive field. Transformer and MLP naturally have global receptive filed, while convolution has local receptive field, e.g., $3 \times 3$. When combining those operators, there is no single optimal DSM that satisfies all scenarios.

The proposed DSMs are visualized in Figure \ref{fig:dsms}. To down-sample based on global cues, we utilize the self-attention mechanism to capture global context, which is missed by the prior art. However, vanilla self-attention directly applies a linear transformation on input $\mathtt{x}$, which is not applicable to down-sample a feature map. To address this issue, we replace the original linear transformation of a query with down-sampling operations. Specifically, for G-DSM, we use $\mathtt{Conv1D}$ with stride 2 to down-sample the query. To note that, there is no direct local context preserved after the transformation of G-DSM. For LG-DSM, we first reshape the flattened token sequences back to the spatial grid and apply $\mathtt{Conv2D}$ with stride 2 to down-sample the query, and then flatten the query back to calculate the attention weights.

Compared with previous works, which mainly try to improve ConvNets, our proposed DSMs are not designed for a specific architecture. Our motivation is that the optimal DSM is not fixed for different operators. For example, the optimal DSM may be L-DSM for ConvNets, but G-DSM for transformer. As thousands of operator combinations will be trained in our NAS process, it is unfeasible to decide which DSM to use by hand. To obtain the optimal architecture, we jointly search DSMs with the General Operators. An interesting result in our experiments is that our searched DSM matches our assumption. In our searched optimal architecture, L-DSM is used between operators with local receptive field while LG-DSM is favored by operators with global receptive field. 

\subsection{UniNet Architecture}

As shown in recent studies, combining different operators \citep{cvt, yuan2021incorporating} can bring performance improvements. Most previous approaches only repeatedly stack the same operation in the whole architecture and use different channels in different stages. These approaches don't permit stage diversity, which we show is crucial for achieving high accuracy.

On the contrary, in our UniNet, the operators are not fixed but searched from the unified search space. We construct our UniNet architecture in a multi-stage fashion, which can be easily transferred to downstream tasks. Between two successive stages, one of our proposed DSMs will be inserted to reduce the spatial dimension. We jointly search the GOP and DSM per stage separately. The GOP could be different for different stages but repeated in one stage, which can greatly reduce the search space size as pointed out before \citep{mnas}. The overall architecture and search space are shown in Figure \ref{fig:backbone}. 

Thanks to our unified form of GOPs, the network size of each stage can be configured with repeat number $\mathtt{r}$, channel size $\mathtt{c}$, and expansion ratio $\mathtt{e}$. To obtain better computation-accuracy trade-off, we jointly search the network size with the GOP and DSM. For GOP, we search for convolution, transformer, MLP and their promising variants, i.e., \{$\mathtt{SA}$, $\mathtt{LSA}$, $\mathtt{Conv}$, $\mathtt{DWConv}$, $\mathtt{MLP}$\}, as mentioned in Section \ref{sec3.2}; for $\mathtt{e}$, we search for \{2, 3, 4, 5, 6\}. 
Following previous work, we search the network size based on a given architecture, e.g. EfficientNet \citep{efficientnet}, and the channels and repeats will be changed according to it. For $\mathtt{c}$ and $\mathtt{r}$, we search for \{0.5, 0.75, 1.0, 1.25, 1.5\} and \{-2, -1, 0, 1, 2\} respectively.
Suppose we partition the network into $K$ stages, and each stage has a sub search space of size $S$. Then the total search space is $S^N$. In our implementation, $K$ is set to 5 and $S$ equals 125. As a result, our search space size is about 10\textsuperscript{16}.

\subsection{Search Algorithm}

We utilize Reinforcement Learning to search for the optimal architecture automatically. Concretely, we follow previous work \citep{fnas} and map an architecture in the search space to a list of tokens, which are determined by a sequence of actions generated by an RNN. The RNN is optimized using PPO algorithm \citep{ppo} by maximizing the expected reward. In our implementation, we simultaneously optimize accuracy and the theoretical computation cost (FLOPs). To handle the multi-objective optimization problem, we use a weighted product customized as \citet{mnas} to approximate Pareto optimal. For one sampled architecture $m$, the reward is formulated as $r(m) = a(m) \times \frac{t}{f(m)}^{\alpha}$, where function $a(m)$ and $f(m)$ return the accuracy and the FLOPs of $m$, $t$ is the target FLOPs, and $\alpha$ is a weight factor that balances the accuracy and computation cost.

During the search process, thousands of combinations of GOPs and DSMs are trained on a proxy task with the same setting, which gives us a fair comparison between those combinations.  When the RNN converges, the top-k architectures with the highest reward will be trained with full setting, and the top-performing one will be kept for model scaling and transferring to other downstream tasks.

\section{Experimental Setup}

To find the optimal architecture in our search space, we directly search on a large dataset, ImageNet. We reserve 50K images from the training set as a validation set. For each sampled architecture, we train it for 5 epochs. After that, we calculate the reward of the architecture with its FLOPs and the accuracy on the validation set. We set the target FLOPs $t$ and weight factor $\alpha$ in reward function to 550M and -0.07 respectively. During the search process, totally 2K models are trained on the proxy task. After that, we fully train the top 5 architectures on ImageNet and preserve the top-performing one for model scaling and transferring to other downstream tasks.

For regular ImageNet training, we mostly follow the training strategy in DeiT \citep{deit}, except that we use small augmentation for small models and heave augmentation for large models as shown by \citet{trainvit}. When the input resolution exceeds $256 \times 256$, we train our models with $256 \times 256$ and finetune on the large resolution for training efficiency. Besides, we also transfer UniNet to other downstream tasks, e.g., object detection on COCO and semantic segmentation on ADE20K.
For COCO training, we use the typical framework Mask R-CNN and train on wildly-used 1x (12 epochs) and 3x (36 epochs) schedules. For ADE20K training, we use the UperNet framework and train with the same setting as \citet{swin}.

\section{Main Results}
\label{main_results}

In this section, we firstly present our searched UniNet architecture. We then show the performance of the scaled UniNets on classification, object detection, and semantic segmentation.

\begin{table}[t]
  \centering
    \begin{tabular}{lc|ccc|c}
    \toprule
    \multirow{2}{*}{Model} & \multirow{2}{*}{Family} & \multirow{2}{*}{\makecell{Input \\ Size}} & \multirow{2}{*}{\makecell{\#FLOPs \\ (G)} } & \multirow{2}{*}{\makecell{\#Params \\ (M)}} & \multirow{2}{*}{\makecell{Top-1 \\ Acc.}} \\
    & & & & & \\
    \midrule
    EffNet-B0 \citep{efficientnet} & C     & 224   & 0.39  & 5.3   & 77.1 \\
    EffNetV2-B0\textsuperscript{$\ddagger$} \citep{effnetv2} & C     & 240   & 0.7   & 7.4   & 78.7 \\
    DeiT-Tiny \citep{deit} & T     & 224   & 1.3   & 5.7   & 72.2 \\
    PVT-Tiny \citep{pvt} & T     & 224   & 1.9   & 13.2  & 75.1 \\
    ConViT-Ti+ \citep{convit} & H     & 224   & 2     & 10    & 76.7 \\
    \rowcolor{LightCyan}
    UniNet-B0 & H     & 160   & 0.56  & 11.9  & 79.1 \\
    \midrule
    EffNet-B2 \citep{efficientnet} & C     & 260   & 1     & 9.2   & 80.1 \\
    EffNetV2-B1\textsuperscript{$\ddagger$} \citep{effnetv2} & C     & 260   & 1.2   & 8.1   & 79.8 \\
    RegNetY-4G \citep{regnet} & C     & 224   & 4     & 20.6  & 80 \\
    DeiT-Small \citep{deit} & T     & 224   & 4.3   & 22    & 79.8 \\
    PVT-Small \citep{pvt} & T     & 224   & 3.8   & 24.5  & 79.8 \\
    \rowcolor{LightCyan}
    UniNet-B1 & H     & 192   & 0.99  & 14    & 80.4 \\
    \midrule
    EffNet-B3 \citep{efficientnet} & C     & 300   & 1.8   & 12    & 81.6 \\
    EffNetV2-B3\textsuperscript{$\ddagger$} \citep{effnetv2} & C     & 300   & 3     & 14    & 82.1 \\
    Swin-T \citep{swin} & T     & 224   & 4.5   & 29    & 81.3 \\
    CvT-13-NAS \citep{cvt} & H     & 224   & 4.1   & 18    & 82.2 \\
    ConViT-B+ \citep{convit} & H     & 224   & 30    & 152   & 82.5 \\
    \rowcolor{LightCyan}
    UniNet-B2 & H     & 224   & 2.4   & 22.5  & 82.7  \\
    \midrule
    EffNet-B4 \citep{efficientnet} & C     & 380   & 4.2   & 19    & 82.9 \\
    NFNet-F0 \citep{nfnet} & C     & 256   & 12.4  & 71.5  & 83.6 \\
    Swin-B \citep{swin} & T     & 224   & 15.4  & 88    & 83.5 \\
    CvT-21 \citep{cvt} & H     & 384   & 24.9  & 32    & 83.3 \\
    \rowcolor{LightCyan}
    UniNet-B3 & H     & 256   & 4.2   & 31.8  & 83.7 \\
    \rowcolor{LightCyan}
    UniNet-B4 & H     & 256   & 9.9   & 73.5  & 84.2 \\
    \midrule
    EffNet-B7 \citep{efficientnet} & C     & 600   & 37    & 66    & 84.3 \\
    EffNetV2-M\textsuperscript{$\ddagger$} \citep{effnetv2} & C     & 480   & 24    & 54    & 85.1 \\
    NFNet-F2 \citep{nfnet} & C     & 352   & 62.6  & 193.8 & 85.1 \\
    BoTNet-T7 \citep{botnet} & T     & 384   & 45.8  & 75.1  & 84.7 \\
    \rowcolor{LightCyan}
    UniNet-B5 & H     & 384   & 23.2  & 73.5  & 85.2 \\
    \bottomrule
  \end{tabular}%
   \vspace{-1em}
  \caption{UniNet performance on ImageNet. All UniNet models are trained with ImageNet dataset with 1.28m images. C, T, and H denotes convolution, transformer, and hybrid architecture respectively. $\ddagger$: all EfficientNetV2 models are trained with progressive learning.}
  \label{tab:sota}%
   \vspace{-1.8em}
\end{table}%

\subsection{The Searched UniNet}

\begin{wraptable}{r}{0.5\textwidth}
    \centering
    \resizebox{0.45\textwidth}{!}{
    \begin{tabular}{c|cc|ccc}
    \toprule
    \multirow{2}{*}{Stage}  & \multicolumn{2}{c}{Operator} & \multicolumn{3}{c}{Network Size} \\
     & GOP & DSM & $\mathtt{e}$ & $\mathtt{c}$ & $\mathtt{r}$ \\
    \midrule
    0     & DWConv & L-DSM & 4     & 48    & 2 \\
    1     & DWConv & L-DSM & 6     & 80    & 4 \\
    2     & DWConv & L-DSM & 3     & 128   & 4 \\
    3     & Transformer & LG-DSM & 2     & 128   & 4 \\
    4     & Transformer & LG-DSM & 5     & 256   & 8 \\
    \bottomrule
    \end{tabular}
    }
    \vspace{-1em}
    \caption{UniNet-B0 architecture. GOP and DSM represent General Operators and down-sampling module respectively. DWConv and Transformer are are described in Section \ref{sec3.2}.}
    \vspace{-1em}
    \label{tab:uninet_b05}
\end{wraptable}

Table \ref{tab:uninet_b05} shows our searched UniNet-B0 architecture. Our searched architecture has the following characteristics: (1) UniNet uses DWConv at early stages, while uses transformer at later stages. (2) UniNet chooses L-DSM to down-sample in the DWConv stage, while uses LG-DSM for transformer.  Surprisingly, the searched operator combination somewhat matches previous empirical finds \citep{earlyconv} or manual designed architecture \citep{container}, which show the effectiveness of our proposed unified architecture search. 

Most previous transformer based architectures usually outperform convolution based architecture in high FLOPs region, but underperform in low FLOPs region.  To prove the effectiveness of our searched UniNet, we scale up and down our search UniNet-B0. The FLOPs number varies from 0.56G to 23.2G, which covers both the mobile and large settings. We utilize the compound scaling \citep{efficientnet} to scale depth, width, and resolution simultaneously. An exception is that we increase image resolution more slowly, which is more efficient as \citet{revisiting} shows.

\subsection{ImageNet Classification Performance}
Table \ref{tab:sota} presents the performance comparison of our searched UniNet and other architectures, including convolution based, transformer based, and hybrid architectures. Our search UniNet has better accuracy and computation efficiency than previous ConvNets, transformer, or hybrid architectures. 

As shown in Table \ref{tab:sota}, under mobile setting, our UniNet-B0 achieves 79.1\% top-1 accuracy with 0.56G FLOPs, outperforming EfficientNetV2-B0 \citep{effnetv2} with less FLOPs. In the middle FLOPs region, our UniNet-B3 achieves 83.7\% top-1 accuracy with 4.2G FLOPs, which outperforms pure convolution based EfficientNet-B4, pure transformer based Swin-B, and hybrid architecture CvT-21. For larger models, our UniNet-B5 achieve 85.2\% with 23.2G FLOPs, outperforming NFNet-F2 and BoTNet-T7 with 63\% and 49\% fewer FLOPs respectively. Figure \ref{fig:sota} further visualizes the comparison with on accuracy and FLOPs. 

\subsection{Object Detection and Semantic Segmentation Performance}

\begin{table}[h]
 \centering
  \begin{tabular}{c|c|cc|cc|cc}
\toprule
\multirow{2}{*}{Backbone} & \multirow{2}{*}{\makecell{\#Params (M) \\ Det/Seg}} & \multicolumn{2}{c}{Mask R-CNN 1x} & \multicolumn{2}{c}{Mask R-CNN 3x} & \multirow{2}{*}{\makecell{UperNet \\ mIoU (\%)}} \\  
 & & AP@box & AP@mask & AP@box & AP@mask &  \\ 
 \midrule
 ResNet18   & 31/ -  & 34.0 & 31.2 & 36.9 & 33.6 & -  \\
 ResNet50   & 44/ -  & 38.0 & 34.4 & 41.0 & 37.1 & -  \\
 PVT-Tiny   & 33/ -  & 36.7 & 35.1 & 39.8 & 37.4 & -  \\
 \rowcolor{LightCyan}
 UniNet-B1 & 28/38 & 40.5 & 37.5 & 44.4 & 40.1 & 42.7 \\
 \midrule
 ResNet101  & 63/86  & 40.4 & 36.4 & 42.8 & 38.5 & 44.9  \\
 PVT-Small  & 44/ -  & 40.4 & 37.8 & 43.0 & 39.9 & -  \\
 Swin-T     & 48/60  & 42.2 & 39.1 & 46.0 & 41.6 & 44.5  \\ 
 \rowcolor{LightCyan}
 UniNet-B3 & 50/59 & 45.6 & 41.6 & 47.9 & 43.3 & 49.0 \\
 \bottomrule
\end{tabular}
\vspace{-1em}
\caption{Object detection, instance segmentation, and semantic segmentation performance on the COCO val2017 and ADE20K val set. All UniNet models
are pretrained on the ImageNet-1K dataset.}
 \label{tab:det_seg_results}
\end{table}

For object detection and semantic segmentation, we pick UniNet-B1 and UniNet-B3 and use them as the feature extractors of detection and segmentation frameworks. We compare our UniNet with other convolution or transformer based architectures. For COCO object detection, we use Mask-RCNN framework and compare the performance under 1x and 3x schedules. For ADE20K semantic segmentation we use UperNet framework and report mIoU (\%) for different architectures under same the training setting.

As shown in Table \ref{tab:det_seg_results}, our searched UniNet consistent outperforming convolution based ResNet \citep{resnet} and transformer based PVT \citep{pvt} or Swin-Transformer \citep{swin}. UniNet-B1 achieves 40.5 AP@box, which is 3.8 points better than PVT-Tiny but with 15\% fewer parameters. UniNet-B4 achieves 45.6 AP@box with 1x schedule and 47.9 AP@box with 3x schedule, which is 3.4 points and 1.9 points better than Swin-T respectively. For ADE20K semantic segmentation, we achieve 49\% mIoU with 59M parameters. Compared with transformer based Swin-T, our UniNet outperforms 4.5\% mIoU with similar parameters. Besides, compared with convolution based ResNet101, we get 4.1\% higher mIoU with 31\% fewer parameters. All the results show the generalization of our searched UniNet. 

\section{Ablative Studies and Analysis}
\label{ablation}

In this section, we study the impact of joint search of General Operations, and discuss the importance of context-aware down-sampling modules (DSMs).

\subsection{Single Operator vs. General Operators}
\begin{table}[h]
    \centering
        \resizebox{0.8\textwidth}{!}{
        \begin{tabular}{c | cc | c}
        \midrule
        Model & \#FLOPs (G) & \#Params (M) & {Top-1 Acc.} \\
        \midrule
        UniNet-B0   & 0.56     & 11.5     & \textbf{79.1} \\
        \midrule
        UniNet-B0 w/ Conv-Only & 0.59     & 11.0     & 77.7 \\
        \midrule
        \end{tabular}
        }
    \vspace{-1em}
    \caption{Performance on ImageNet of UniNet with different search settings. Conv-Only represents search UniNet with convolution operator only.}
    \label{tab:with_conv}
\end{table}

Previous work \citep{mnas} focus on the network size search, which uses single operator, convolution, as the main feature extractor. In comparison, we jointly search the combination of different General Operators (GOPs), i.e., convolution, transformer, MLP, and their promising variants. To verify the importance of GOPs, we remove MLP and transformer and their variants from our search space, and re-run the search experiment under the same setting. After search, we fully train the top-5 architectures with highest reward on ImageNet, and report the best performance. 

As shown in Table \ref{tab:with_conv}, our joint search of GOPs outperforms Conv-Only search by a large margin. The result verifies the effectiveness of our joint search of GOPs, which can combine the characteristics of different operators. To note that, the DSMs are also searched in the Conv-Only search experiment. However, the top-performing architecture chooses L-DSM in all its stages.

\subsection{Fixed vs. Context-Aware Down-Sampling Module}

\begin{table}[h]
  \centering
    \begin{minipage}[t]{0.5\linewidth}
        \centering
        \resizebox{\textwidth}{!}{
        \begin{tabular}{c | cc | c}
        \toprule
        \multirow{2}{*}{Model} & \multirow{2}{*}{\makecell{\#FLOPs \\ (G)}} & \multirow{2}{*}{ \makecell{\#Params \\ (M)}} & \multirow{2}{*}{\makecell{Top-1 \\ Acc.}} \\
        &  &  & \\
        \midrule
        UniNet            & 0.56     & 11.5     & \textbf{79.1} \\
        w/ L-DSM   & 0.54     & 11.3     & 78.5 \\
        w/ G-DSM   & 0.77     & 12.7     & 76.8 \\
        w/ LG-DSM  & 0.72     & 14.1     & 78.9 \\
        \bottomrule
        \end{tabular}
        }
        \vspace{-1em}
        \caption{Performance on ImageNet of UniNet with different DSMs. To note that, the result of traditional strided-conv based down-sampling module is shown in row 2. }
        \label{tab:dsm}
    \end{minipage}
    \hspace{0.5em}
    \begin{minipage}[t]{0.45\linewidth}
        \centering
        \resizebox{\textwidth}{!}{
        \begin{tabular}{c|cc|c}
        \toprule
        \multirow{2}{*}{Model} & \multirow{2}{*}{\makecell{\#FLOPs \\ (G)}} & \multirow{2}{*}{ \makecell{\#Params \\ (M)}} & \multirow{2}{*}{\makecell{Top-1 \\ Acc.}} \\
        &  &  & \\
        \midrule
        PVT-Tiny         & 1.9 & 13.2 & 75.1 \\ 
        w/ DSM  & 2.0 & 14.3 & \textbf{77.5} \\
        \midrule
        Swin-T        & 4.5 & 29.0 & 81.2 \\
        w/ DSM & 4.7 & 30.0 & \textbf{81.6} \\
        \bottomrule
        \end{tabular}
        }
        \vspace{-1em}
        \caption{Performance comparison on ImageNet of different backbones when equipped with our proposed DSMs. }
        \label{tab:transfer_dsm}
    \end{minipage}
\end{table}

When combining different operators into a unified network, the traditional down-sampling module, which is mainly implemented with strided-conv or pooling, could be sub-optimal. To verify the effectiveness of our proposed context-aware DSMs, we replace the DSMs of our search UniNet with one fixed DSM, and compare their performance under the same training setting.

As shown in Table \ref{tab:dsm}, our searched UniNet consistently outperforms its variants that use single fixed DSM in all stages. Although we see that using G-DSM or LG-DSM in all stages brings more computation and parameters, the performance does not become better. The result emphasizes the importance of our joint search of GOPs and DSMs.

Besides, we transfer our proposed DSMs to other popular transformer based architectures, e.g. Swin-Transformer \citep{swin} and PVT \citep{pvt}. Both Swin and PVT have 4 stages. Borrowing the result from our searched UniNet, we use L-DSM for the first two stages while LG-DSM for the latter two stages. As shown in Table \ref{tab:transfer_dsm}, our proposed DSMs improve PVT-Tiny and Swin-T for 2.4\% and 0.4\% respectively. To note that, PVT uses a strided-conv to down-sample. As discussed in Section \ref{sec:dsm}, it is harmful to the main operator in PVT, transformer, which has global receptive field. On the contrary, our proposed DSMs are able to down-sample based on both local and global context, and can greatly improve the performance.



\section{Conclusion}
\label{conclusion}

In this paper, we propose to jointly search the combination of convolution, transformer, and MLP. We empirically identify that the widely-used strided convolution or pooling based down-sampling modules become the performance bottlenecks when the operators are combined to form a network. To further improve the performance, we propose context-aware down-sampling modules and jointly search them with all operators. Our searched UniNet outperforms state-of-the-art pure convolution-based architecture, EfficientNet, and pure transformer-based architecture, Swin-Transformer, on ImageNet classification, COCO object detection, and ADE20K semantic segmentation.

\bibliography{iclr2022_conference}

\begin{thebibliography}{32}
\providecommand{\natexlab}[1]{#1}
\providecommand{\url}[1]{\texttt{#1}}
\expandafter\ifx\csname urlstyle\endcsname\relax
  \providecommand{\doi}[1]{doi: #1}\else
  \providecommand{\doi}{doi: \begingroup \urlstyle{rm}\Url}\fi

\bibitem[Bello et~al.(2021)Bello, Fedus, Du, Cubuk, Srinivas, Lin, Shlens, and
  Zoph]{revisiting}
Irwan Bello, William Fedus, Xianzhi Du, Ekin~D Cubuk, Aravind Srinivas,
  Tsung-Yi Lin, Jonathon Shlens, and Barret Zoph.
\newblock Revisiting resnets: Improved training and scaling strategies.
\newblock \emph{arXiv preprint arXiv:2103.07579}, 2021.

\bibitem[Brock et~al.(2021)Brock, De, Smith, and Simonyan]{nfnet}
Andrew Brock, Soham De, Samuel~L Smith, and Karen Simonyan.
\newblock High-performance large-scale image recognition without normalization.
\newblock \emph{arXiv preprint arXiv:2102.06171}, 2021.

\bibitem[Chollet(2017)]{xception}
Fran{\c{c}}ois Chollet.
\newblock Xception: Deep learning with depthwise separable convolutions.
\newblock In \emph{Proceedings of the IEEE conference on computer vision and
  pattern recognition}, pp.\  1251--1258, 2017.

\bibitem[d'Ascoli et~al.(2021)d'Ascoli, Touvron, Leavitt, Morcos, Biroli, and
  Sagun]{convit}
St{\'e}phane d'Ascoli, Hugo Touvron, Matthew Leavitt, Ari Morcos, Giulio
  Biroli, and Levent Sagun.
\newblock Convit: Improving vision transformers with soft convolutional
  inductive biases.
\newblock \emph{arXiv preprint arXiv:2103.10697}, 2021.

\bibitem[Deng et~al.(2009)Deng, Dong, Socher, Li, Li, and Fei-Fei]{imagenet}
Jia Deng, Wei Dong, Richard Socher, Li-Jia Li, Kai Li, and Li~Fei-Fei.
\newblock Imagenet: A large-scale hierarchical image database.
\newblock In \emph{2009 IEEE conference on computer vision and pattern
  recognition}, pp.\  248--255. Ieee, 2009.

\bibitem[Dosovitskiy et~al.(2020)Dosovitskiy, Beyer, Kolesnikov, Weissenborn,
  Zhai, Unterthiner, Dehghani, Minderer, Heigold, Gelly, et~al.]{vit}
Alexey Dosovitskiy, Lucas Beyer, Alexander Kolesnikov, Dirk Weissenborn,
  Xiaohua Zhai, Thomas Unterthiner, Mostafa Dehghani, Matthias Minderer, Georg
  Heigold, Sylvain Gelly, et~al.
\newblock An image is worth 16x16 words: Transformers for image recognition at
  scale.
\newblock \emph{arXiv preprint arXiv:2010.11929}, 2020.

\bibitem[Gao et~al.(2021)Gao, Lu, Li, Mottaghi, and Kembhavi]{container}
Peng Gao, Jiasen Lu, Hongsheng Li, Roozbeh Mottaghi, and Aniruddha Kembhavi.
\newblock Container: Context aggregation network.
\newblock \emph{arXiv preprint arXiv:2106.01401}, 2021.

\bibitem[Gao et~al.(2019)Gao, Wang, and Wu]{lip}
Ziteng Gao, Limin Wang, and Gangshan Wu.
\newblock Lip: Local importance-based pooling.
\newblock In \emph{Proceedings of the IEEE/CVF International Conference on
  Computer Vision}, pp.\  3355--3364, 2019.

\bibitem[He et~al.(2016)He, Zhang, Ren, and Sun]{resnet}
Kaiming He, Xiangyu Zhang, Shaoqing Ren, and Jian Sun.
\newblock Deep residual learning for image recognition.
\newblock In \emph{Proceedings of the IEEE conference on computer vision and
  pattern recognition}, pp.\  770--778, 2016.

\bibitem[Lin et~al.(2014)Lin, Maire, Belongie, Hays, Perona, Ramanan,
  Doll{\'a}r, and Zitnick]{coco}
Tsung-Yi Lin, Michael Maire, Serge Belongie, James Hays, Pietro Perona, Deva
  Ramanan, Piotr Doll{\'a}r, and C~Lawrence Zitnick.
\newblock Microsoft coco: Common objects in context.
\newblock In \emph{European conference on computer vision}, pp.\  740--755.
  Springer, 2014.

\bibitem[Liu et~al.(2021{\natexlab{a}})Liu, Zhang, Sun, Liu, Song, Liu, and
  Li]{fnas}
Jihao Liu, Ming Zhang, Yangting Sun, Boxiao Liu, Guanglu Song, Yu~Liu, and
  Hongsheng Li.
\newblock Fnas: Uncertainty-aware fast neural architecture search.
\newblock \emph{arXiv preprint arXiv:2105.11694}, 2021{\natexlab{a}}.

\bibitem[Liu et~al.(2021{\natexlab{b}})Liu, Lin, Cao, Hu, Wei, Zhang, Lin, and
  Guo]{swin}
Ze~Liu, Yutong Lin, Yue Cao, Han Hu, Yixuan Wei, Zheng Zhang, Stephen Lin, and
  Baining Guo.
\newblock Swin transformer: Hierarchical vision transformer using shifted
  windows.
\newblock \emph{arXiv preprint arXiv:2103.14030}, 2021{\natexlab{b}}.

\bibitem[Radosavovic et~al.(2020)Radosavovic, Kosaraju, Girshick, He, and
  Doll{\'a}r]{regnet}
Ilija Radosavovic, Raj~Prateek Kosaraju, Ross Girshick, Kaiming He, and Piotr
  Doll{\'a}r.
\newblock Designing network design spaces.
\newblock In \emph{Proceedings of the IEEE/CVF Conference on Computer Vision
  and Pattern Recognition}, pp.\  10428--10436, 2020.

\bibitem[Saeedan et~al.(2018)Saeedan, Weber, Goesele, and Roth]{dpp}
Faraz Saeedan, Nicolas Weber, Michael Goesele, and Stefan Roth.
\newblock Detail-preserving pooling in deep networks.
\newblock In \emph{Proceedings of the IEEE Conference on Computer Vision and
  Pattern Recognition}, pp.\  9108--9116, 2018.

\bibitem[Sandler et~al.(2018)Sandler, Howard, Zhu, Zhmoginov, and Chen]{mbv2}
Mark Sandler, Andrew Howard, Menglong Zhu, Andrey Zhmoginov, and Liang-Chieh
  Chen.
\newblock Mobilenetv2: Inverted residuals and linear bottlenecks.
\newblock In \emph{Proceedings of the IEEE conference on computer vision and
  pattern recognition}, pp.\  4510--4520, 2018.

\bibitem[Schulman et~al.(2017)Schulman, Wolski, Dhariwal, Radford, and
  Klimov]{ppo}
John Schulman, Filip Wolski, Prafulla Dhariwal, Alec Radford, and Oleg Klimov.
\newblock Proximal policy optimization algorithms.
\newblock \emph{arXiv preprint arXiv:1707.06347}, 2017.

\bibitem[Srinivas et~al.(2021)Srinivas, Lin, Parmar, Shlens, Abbeel, and
  Vaswani]{botnet}
Aravind Srinivas, Tsung-Yi Lin, Niki Parmar, Jonathon Shlens, Pieter Abbeel,
  and Ashish Vaswani.
\newblock Bottleneck transformers for visual recognition.
\newblock In \emph{Proceedings of the IEEE/CVF Conference on Computer Vision
  and Pattern Recognition}, pp.\  16519--16529, 2021.

\bibitem[Steiner et~al.(2021)Steiner, Kolesnikov, Zhai, Wightman, Uszkoreit,
  and Beyer]{trainvit}
Andreas Steiner, Alexander Kolesnikov, Xiaohua Zhai, Ross Wightman, Jakob
  Uszkoreit, and Lucas Beyer.
\newblock How to train your vit? data, augmentation, and regularization in
  vision transformers.
\newblock \emph{arXiv preprint arXiv:2106.10270}, 2021.

\bibitem[Szegedy et~al.(2015)Szegedy, Liu, Jia, Sermanet, Reed, Anguelov,
  Erhan, Vanhoucke, and Rabinovich]{inception}
Christian Szegedy, Wei Liu, Yangqing Jia, Pierre Sermanet, Scott Reed, Dragomir
  Anguelov, Dumitru Erhan, Vincent Vanhoucke, and Andrew Rabinovich.
\newblock Going deeper with convolutions.
\newblock In \emph{Proceedings of the IEEE conference on computer vision and
  pattern recognition}, pp.\  1--9, 2015.

\bibitem[Tan \& Le(2019)Tan and Le]{efficientnet}
Mingxing Tan and Quoc Le.
\newblock Efficientnet: Rethinking model scaling for convolutional neural
  networks.
\newblock In \emph{International Conference on Machine Learning}, pp.\
  6105--6114. PMLR, 2019.

\bibitem[Tan \& Le(2021)Tan and Le]{effnetv2}
Mingxing Tan and Quoc~V Le.
\newblock Efficientnetv2: Smaller models and faster training.
\newblock \emph{arXiv preprint arXiv:2104.00298}, 2021.

\bibitem[Tan et~al.(2019)Tan, Chen, Pang, Vasudevan, Sandler, Howard, and
  Le]{mnas}
Mingxing Tan, Bo~Chen, Ruoming Pang, Vijay Vasudevan, Mark Sandler, Andrew
  Howard, and Quoc~V Le.
\newblock Mnasnet: Platform-aware neural architecture search for mobile.
\newblock In \emph{Proceedings of the IEEE/CVF Conference on Computer Vision
  and Pattern Recognition}, pp.\  2820--2828, 2019.

\bibitem[Tolstikhin et~al.(2021)Tolstikhin, Houlsby, Kolesnikov, Beyer, Zhai,
  Unterthiner, Yung, Keysers, Uszkoreit, Lucic, et~al.]{mixer}
Ilya Tolstikhin, Neil Houlsby, Alexander Kolesnikov, Lucas Beyer, Xiaohua Zhai,
  Thomas Unterthiner, Jessica Yung, Daniel Keysers, Jakob Uszkoreit, Mario
  Lucic, et~al.
\newblock Mlp-mixer: An all-mlp architecture for vision.
\newblock \emph{arXiv preprint arXiv:2105.01601}, 2021.

\bibitem[Touvron et~al.(2021{\natexlab{a}})Touvron, Bojanowski, Caron, Cord,
  El-Nouby, Grave, Joulin, Synnaeve, Verbeek, and J{\'e}gou]{resmlp}
Hugo Touvron, Piotr Bojanowski, Mathilde Caron, Matthieu Cord, Alaaeldin
  El-Nouby, Edouard Grave, Armand Joulin, Gabriel Synnaeve, Jakob Verbeek, and
  Herv{\'e} J{\'e}gou.
\newblock Resmlp: Feedforward networks for image classification with
  data-efficient training.
\newblock \emph{arXiv preprint arXiv:2105.03404}, 2021{\natexlab{a}}.

\bibitem[Touvron et~al.(2021{\natexlab{b}})Touvron, Cord, Douze, Massa,
  Sablayrolles, and J{\'e}gou]{deit}
Hugo Touvron, Matthieu Cord, Matthijs Douze, Francisco Massa, Alexandre
  Sablayrolles, and Herv{\'e} J{\'e}gou.
\newblock Training data-efficient image transformers \& distillation through
  attention.
\newblock In \emph{International Conference on Machine Learning}, pp.\
  10347--10357. PMLR, 2021{\natexlab{b}}.

\bibitem[Vaswani et~al.(2017)Vaswani, Shazeer, Parmar, Uszkoreit, Jones, Gomez,
  Kaiser, and Polosukhin]{attention}
Ashish Vaswani, Noam Shazeer, Niki Parmar, Jakob Uszkoreit, Llion Jones,
  Aidan~N Gomez, {\L}ukasz Kaiser, and Illia Polosukhin.
\newblock Attention is all you need.
\newblock In \emph{Advances in neural information processing systems}, pp.\
  5998--6008, 2017.

\bibitem[Wang et~al.(2021{\natexlab{a}})Wang, Chen, Xu, Liu, Loy, and
  Lin]{carafe++}
Jiaqi Wang, Kai Chen, Rui Xu, Ziwei Liu, Chen~Change Loy, and Dahua Lin.
\newblock Carafe++: Unified content-aware reassembly of features.
\newblock \emph{IEEE Transactions on Pattern Analysis and Machine
  Intelligence}, 2021{\natexlab{a}}.

\bibitem[Wang et~al.(2021{\natexlab{b}})Wang, Xie, Li, Fan, Song, Liang, Lu,
  Luo, and Shao]{pvt}
Wenhai Wang, Enze Xie, Xiang Li, Deng-Ping Fan, Kaitao Song, Ding Liang, Tong
  Lu, Ping Luo, and Ling Shao.
\newblock Pyramid vision transformer: A versatile backbone for dense prediction
  without convolutions.
\newblock \emph{arXiv preprint arXiv:2102.12122}, 2021{\natexlab{b}}.

\bibitem[Wu et~al.(2021)Wu, Xiao, Codella, Liu, Dai, Yuan, and Zhang]{cvt}
Haiping Wu, Bin Xiao, Noel Codella, Mengchen Liu, Xiyang Dai, Lu~Yuan, and Lei
  Zhang.
\newblock Cvt: Introducing convolutions to vision transformers.
\newblock \emph{arXiv preprint arXiv:2103.15808}, 2021.

\bibitem[Xiao et~al.(2021)Xiao, Singh, Mintun, Darrell, Doll{\'a}r, and
  Girshick]{earlyconv}
Tete Xiao, Mannat Singh, Eric Mintun, Trevor Darrell, Piotr Doll{\'a}r, and
  Ross Girshick.
\newblock Early convolutions help transformers see better.
\newblock \emph{arXiv preprint arXiv:2106.14881}, 2021.

\bibitem[Yuan et~al.(2021)Yuan, Guo, Liu, Zhou, Yu, and
  Wu]{yuan2021incorporating}
Kun Yuan, Shaopeng Guo, Ziwei Liu, Aojun Zhou, Fengwei Yu, and Wei Wu.
\newblock Incorporating convolution designs into visual transformers.
\newblock \emph{arXiv preprint arXiv:2103.11816}, 2021.

\bibitem[Zhang(2019)]{zhang2019shiftinvar}
Richard Zhang.
\newblock Making convolutional networks shift-invariant again.
\newblock In \emph{ICML}, 2019.

\end{thebibliography}
\bibliographystyle{iclr2022_conference}


\end{document}